\pdfoutput=1
\documentclass[11pt]{article}

\usepackage{emnlp2021}

\usepackage{times}
\usepackage{latexsym}
\usepackage{tabularx}
\usepackage{graphicx}
\usepackage{subfig}
\usepackage[T1]{fontenc}
\usepackage[utf8]{inputenc}

\usepackage{microtype}

\title{AEDA: An Easier Data Augmentation Technique for Text Classification}

\author{Akbar Karimi \hspace{5mm} Leonardo Rossi \hspace{5mm} Andrea Prati \\
   IMP Lab, University of Parma, Italy \\
  \texttt{\{akbar.karimi, leonardo.rossi, andrea.prati\}@unipr.it} \\
  }

\begin{document}
\maketitle
\begin{abstract}
This paper proposes \textbf{AEDA} (\textbf{A}n \textbf{E}asier \textbf{D}ata \textbf{A}ugmentation) technique to help improve the performance on text classification tasks. AEDA includes only random insertion of punctuation marks into the original text.
This is an easier technique to implement for data augmentation than EDA method \citep{wei2019eda} with which we compare our results. In addition, it keeps the order of the words while changing their positions in the sentence leading to a better generalized performance. Furthermore, the deletion operation in EDA can cause loss of information which, in turn, misleads the network, whereas AEDA preserves all the input information. Following the baseline, we perform experiments on five different datasets for text classification. We show that using the AEDA-augmented data for training, the models show superior performance compared to using the EDA-augmented data in all five datasets. The source code is available for further study and reproduction of the results\footnote{\url{https://github.com/akkarimi/aeda_nlp}}.
\end{abstract}

\section{Introduction}

Text classification is a major area of study in natural language processing (NLP) with numerous applications such as sentiment analysis, toxicity detection, and question answering, to name but a few. In order to build text classifiers that perform well, the training data need to be large enough so that the model can generalize to the unseen data. However, for many machine learning (ML) applications and domains, there do not exist sufficient labeled data for training. In this situation, data augmentation (DA) can provide a solution and help improve the performance of ML systems \citep{ragni2014data, fadaee2017data, ding2020daga}. 
DA can be carried out in many different ways such as by modifying elements of the input sequence, namely word substitution, deletion, and insertion \citep{wei2019eda, zhang2015character}, and back-translation \citep{sennrich2016improving}. It can also be performed by noise injection in the input sequence \citep{xie2019data} or in the embedding space utilizing a deep language model \citep{jiao2020tinybert, karimi2020adversarial, garg2020bae}. 

\begin{figure}[t]
\centering
    \includegraphics[scale=0.3]{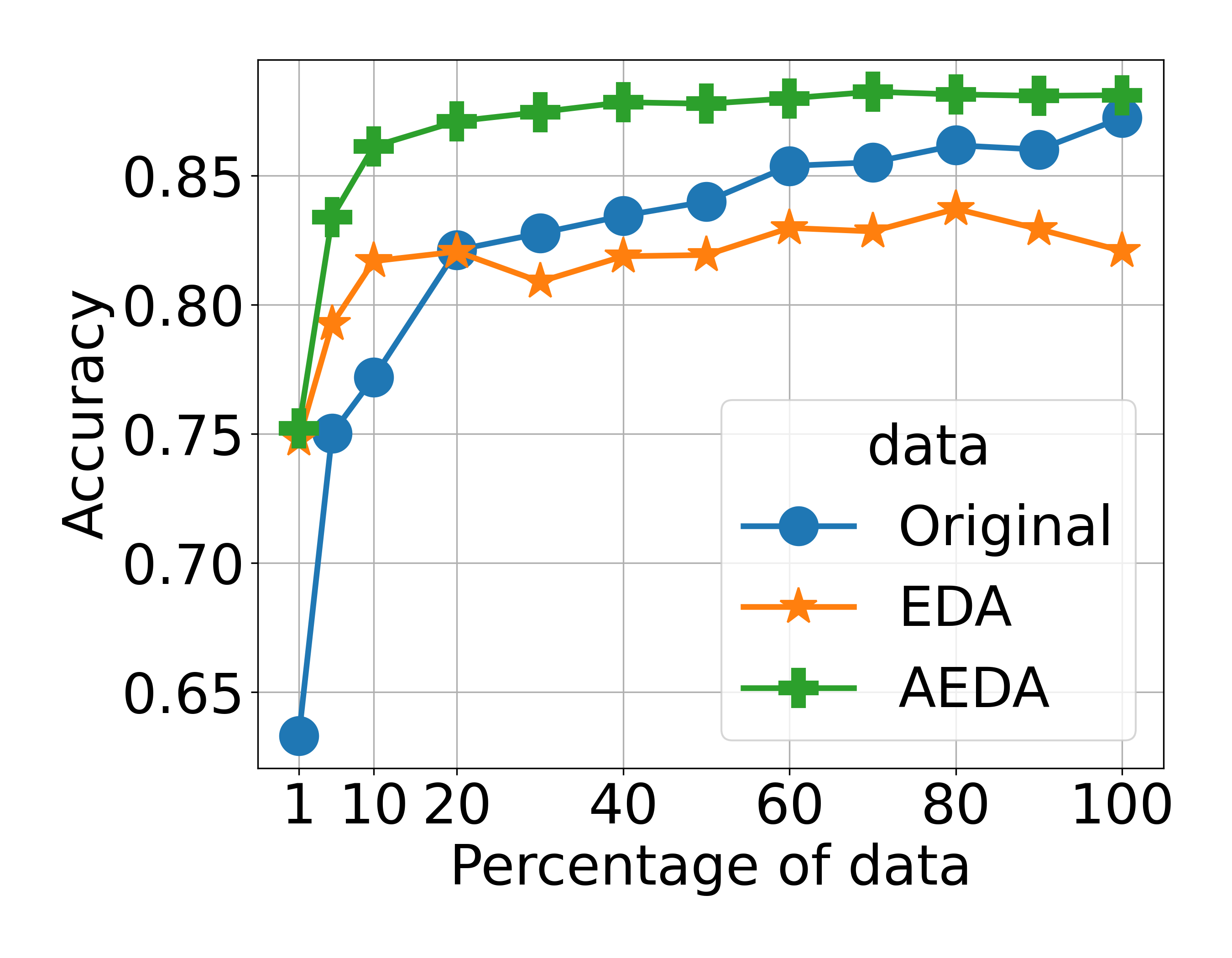}
    \caption{Average performance of the generated data using our proposed augmentation method (AEDA) compared with that of the original and EDA-generated data on five text classification tasks. Using both EDA and AEDA, we added 9 augmented sentences to the original training set to train the models. For each task, we ran the models with 5 different seed numbers and took the average score.}
    \label{fig:average_f1}
\end{figure}

Using a deep language model to do DA can be complicated, while word replacement techniques with the help of a word thesaurus, even though a simple method, risks information loss due to the operations such as deletion and substitution. These operations can even result in changing the label of the input sequence \citep{kumar2020data}, thus misleading the network. 

To address these problems, we propose an extremely simple yet effective approach called AEDA (An Easier Data Augmentation) which includes only the insertion of various punctuation marks into the input sequence. AEDA preserves all the input information and does not mislead the network since it keeps the word order intact while changing their positions in that the words are shifted to the right. Our extensive experiments show that AEDA helps the models avoid overfitting (Figure \ref{fig:average_f1}). 

\section{Related Work}

Although the textual content is always increasing, data augmentation is still a highly active area of research since for machine learning applications, especially the new ones, the initial annotated data are usually small. As a result, researchers are constantly coming up with innovative ideas to create new data from the available content. 

Some have experimented at the input sequence level performing operations on words.
For example, to improve machine translation quality, \citet{fadaee2017data} utilize substitution of common words with rare ones, thus providing more context for the rare words, while \citet{sennrich2016improving} use back-translation where automatically translated data along with the original human-translated data are employed to train a neural machine translation system. \citet{wang2015s} replaces words with their synonyms for classifying tweets. Similarly,
\citet{andreas2020good} replace sentence fragments from common categories with each other in order to produce new sentences.

Others have opted for using pre-trained language models such as BERT \citep{devlin2019bert}. \citet{kobayashi2018contextual} utilizes contextual augmentation, replacing the words with the prediction of a bidirectional language model at a desired position in the sentence. \citet{hu2019learning} and \citet{liu2020data} utilize reinforcement learning with a conditional language model which is carried out by attaching the correct label to the input sequence when training \citep{wu2019conditional}. Working with Transformer model \citep{vaswani2017attention}, \citet{sun2020mixup} propose Mix-Transformer where two input sentences and their corresponding labels are linearly interpolated to create new samples. 

\citet{xie2019data} make use of data noising which can be considered similar to our work with the difference that they replace words choosing from the unigram frequency distribution or insert the underscore character as a placeholder, whereas we insert punctuation characters which usually occur in sentences. 
The related works mostly use some auxiliary data or a complicated language model to produce augmented data. Conversely, our method is extremely simple to implement and does not need any extra data. In addition, it shows superior performance to EDA in both simple models such as RNNs and CNNs and deep models such as BERT. 

\section{AEDA Augmentation}

In order to insert the punctuation marks, we randomly choose a number between 1 and one-third of the length of the sequence which indicates how many insertions will be carried out. The reason is that we want to ensure there is at least on inserted mark and at the same time we do not want to insert too many punctuation marks as too much noise might have a negative effect on the model, although this effect can be investigated in future work. Then, positions in the sequence are also specified in random as many as the selected number in the previous step. In the end, for each chosen position, a punctuation mark is picked randomly from the six punctuation marks in \{".", ";", "?", ":", "!", ","\}. Table \ref{tab:examples}, in Supplementary Material, shows example augmentations by the AEDA technique.

\section{Experimental Setup}
Since we compare our proposed method with \citet{wei2019eda}, we used the same codebase as theirs with no changes in the implementation of the models. We executed the code using a GeForce RTX 2070 GPU with 8 GB of memory. 

\subsection{Datasets}\label{stats}
We experiment with the same five datasets as our baseline. They include \textbf{SST-2} \citep{socher2013parsing}
Standford Sentiment Treebank, \textbf{CR} \citep{hu2004mining, ding2008holistic, liu2015automated}
Customer Reviews Dataset, \textbf{SUBJ} \citep{pang2004sentimental}
Subjectivity/Objectivity Dataset, \textbf{TREC} \citep{li2002learning}
Question Classification Dataset, and \textbf{PC} \citep{ganapathibhotla2008mining}
Pros and Cons Dataset. Table \ref{tab:dataset}, in Supplementary Material, shows the statistics of the utilized datasets.

The train and test sets utilized for the experiments for these datasets were not made available by the baseline. Therefore, after collecting them, we shuffled and divided them into train and test sets with almost the same size as the ones reported by the baseline. For the CR dataset, we combined all the reviews from the three cited sources. The annotations included multiple target sentiments for each sentence. Therefore, to convert them into binary classes, we considered a sentence positive if there was no negative sentiment and negative if there was no positive sentiment. We will make our datasets available along with the source code. 

\subsection{Models}
To be consistent as well as for a fair comparison of the effects of EDA- and AEDA-augmented data, we used the same Recurrent Neural Network (RNN) \citep{liu2016recurrent} and Convolutional Neural Network (CNN) \citep{kim2014convolutional} as implemented in the baseline. 

\section{Results}
To evaluate the quality of augmented sentences, we performed experiments using the data augmented by both EDA and AEDA as well as the original data. For the results reported in Table \ref{tab:resultsrnncnn}, we added 16 augmentations and for the ones in Figure \ref{fig:allfigures}, 9 augmentations to be consistent with the baseline. All experiments were repeated with 5 different seed numbers and the average scores are reported.

\subsection{AEDA Outperforms EDA}
The results of the experiments with 500, 2000, 5000 and full dataset sizes for training are reported in Table \ref{tab:resultsrnncnn}. We can see that in some small datasets, EDA improves the results while for bigger ones it has a negative effect on the performance of the models. Conversely, AEDA gives a performance boost on all datasets, showing greater boosts for smaller ones. For instance, with 500 sentences, the average absolute improvement is 3.2\% while for full dataset it is 0.5\%.  
The reason why EDA does not perform well can be attributed to the operations such as deletion and substitution which insert more misleading information to the network as the number of augmentations grows. In contrast, AEDA keeps the original information in all augmentations. 

\subsection{Trend on Training Set Sizes}
\begin{figure*}
\subfloat[SST-2]{\includegraphics[scale=0.14]{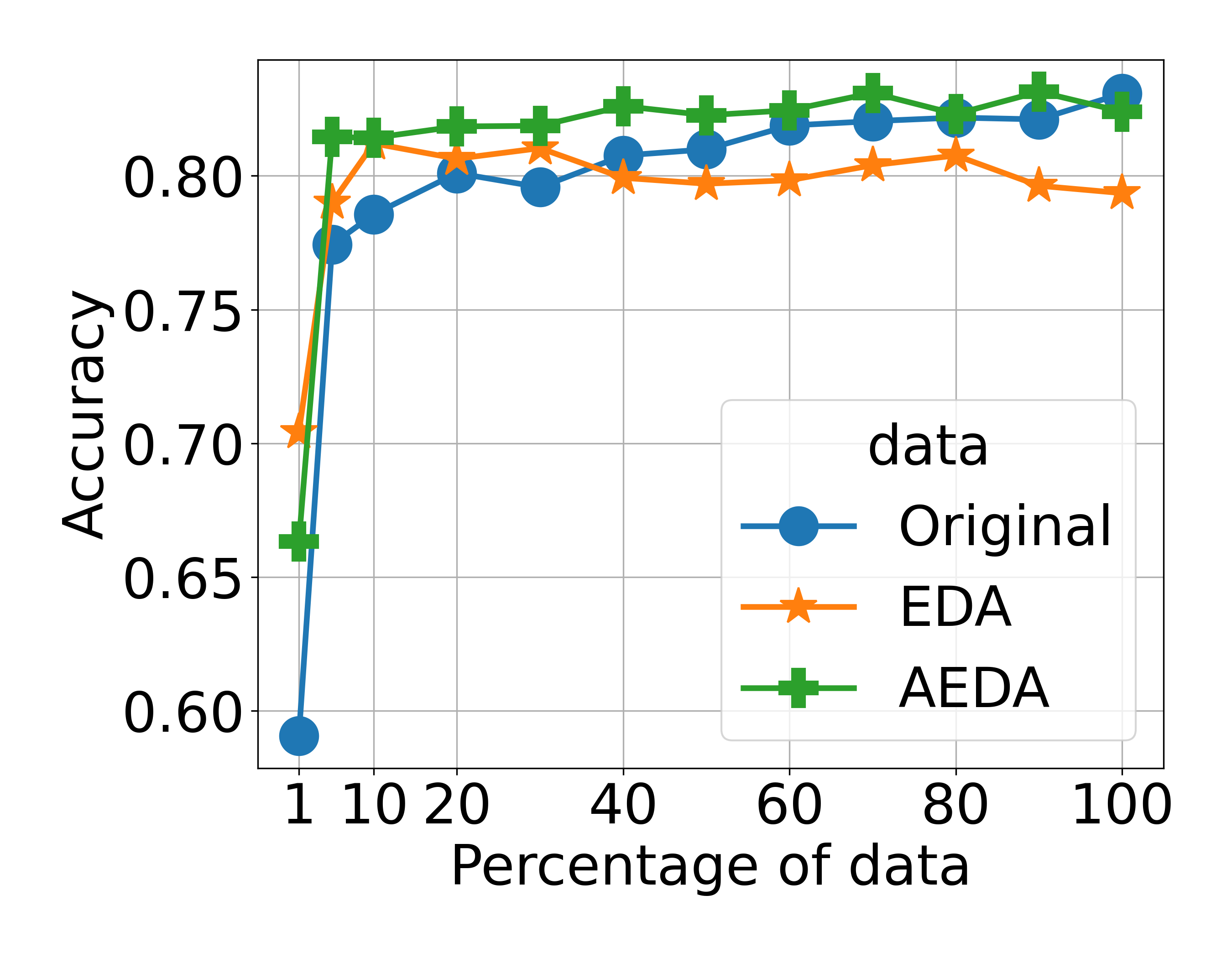}}
\subfloat[CR]{\includegraphics[scale=0.14]{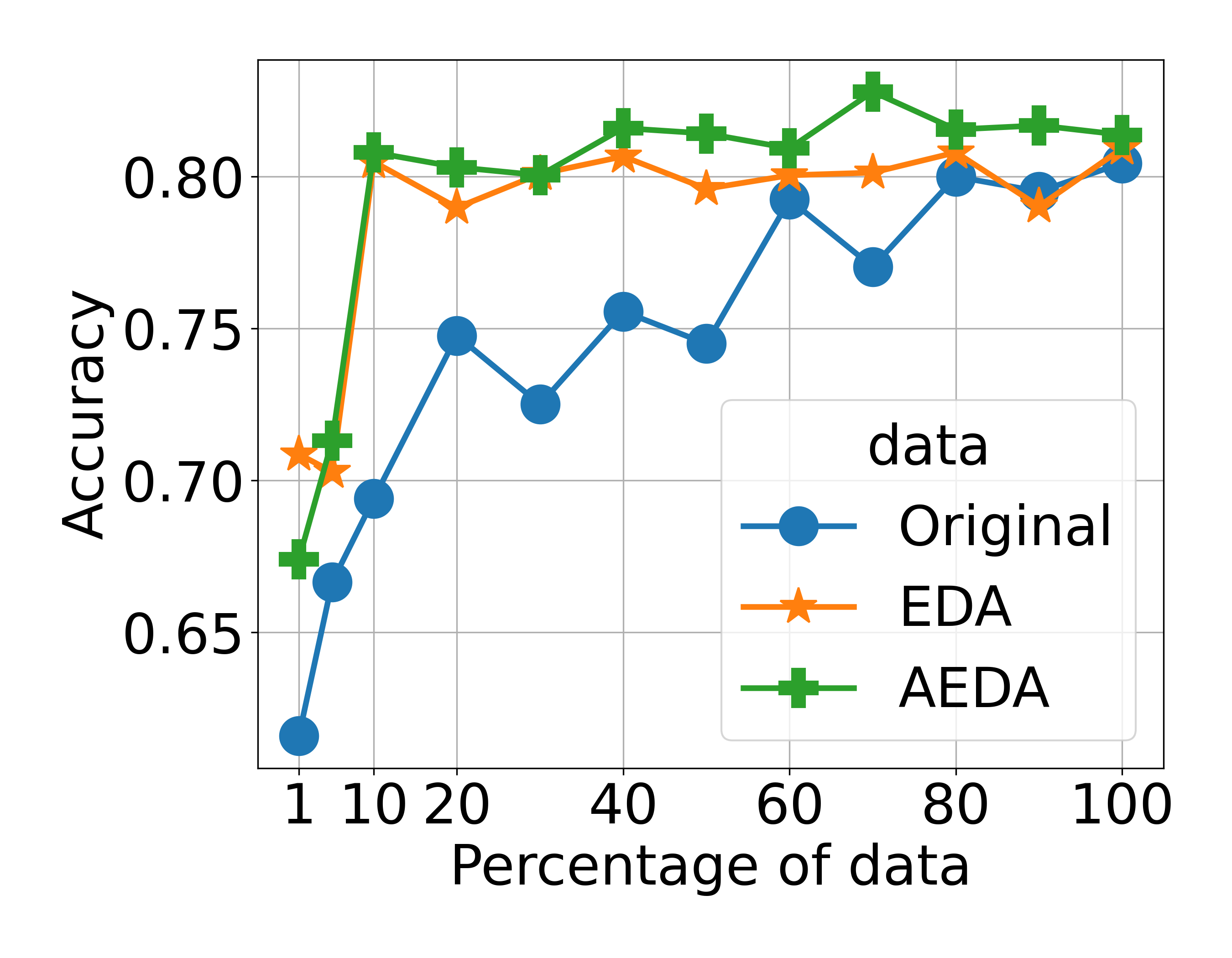}}
\subfloat[SUBJ]{\includegraphics[scale=0.14]{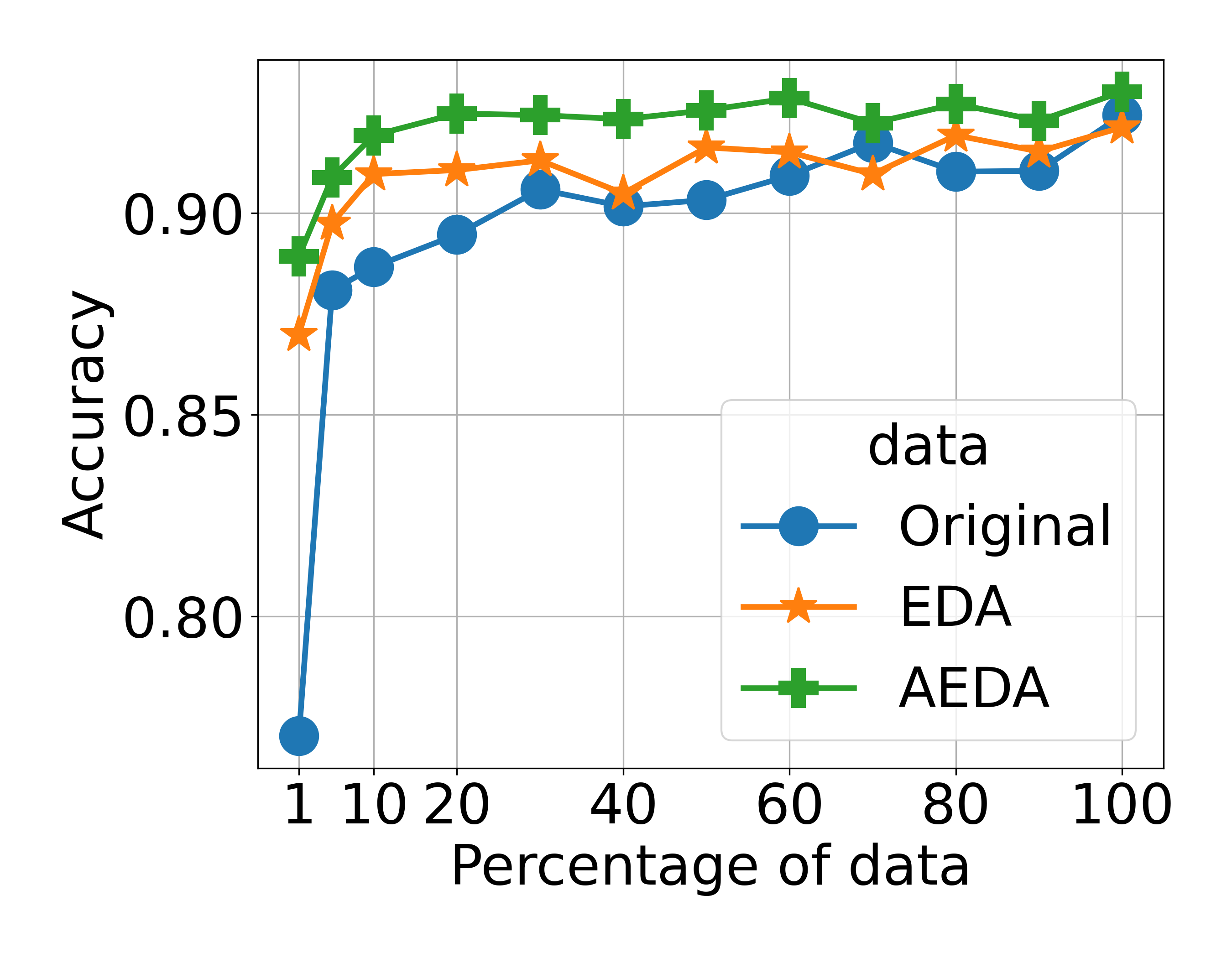}}
\centering
\subfloat[TREC]{\includegraphics[scale=0.14]{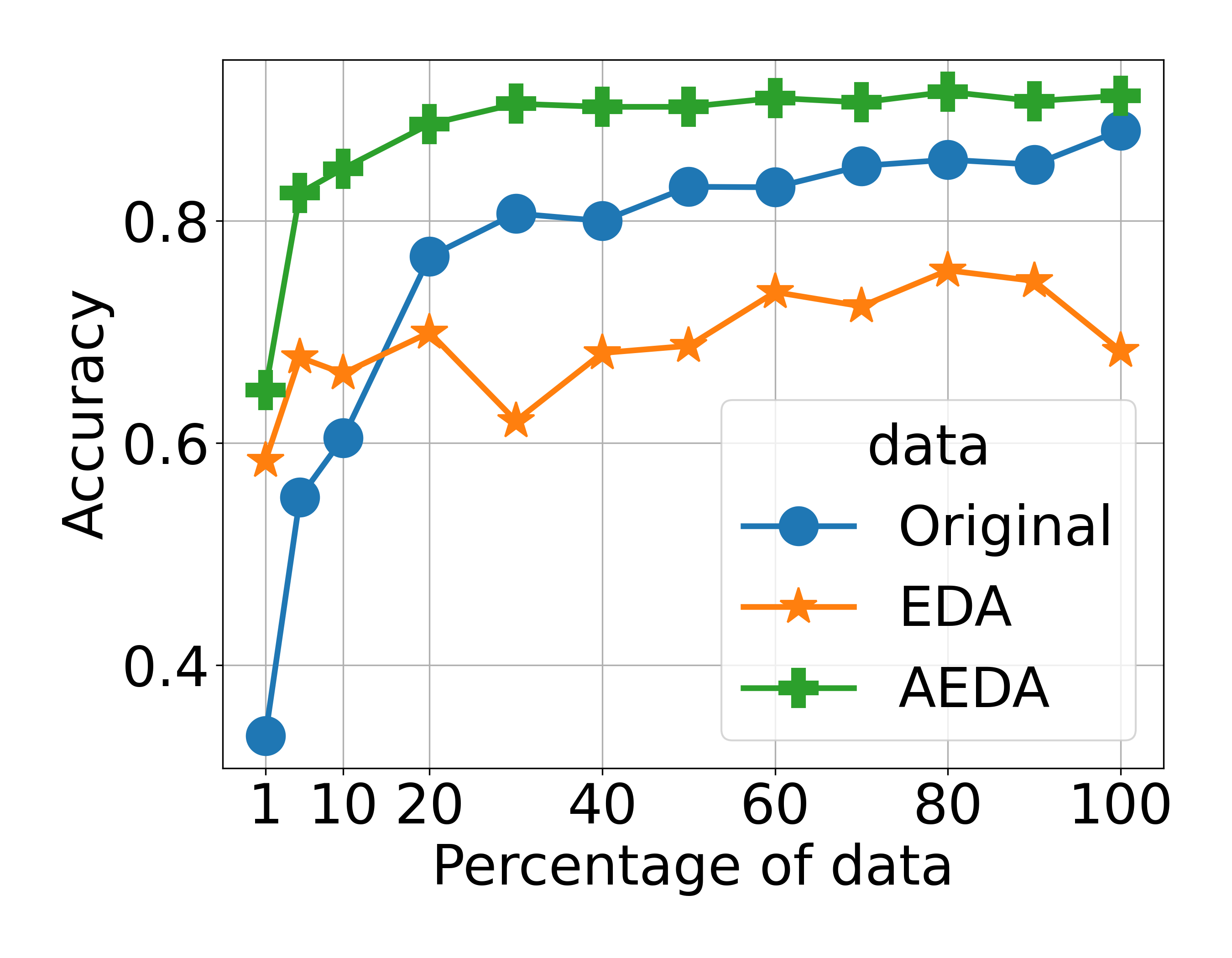}}
\subfloat[PC]{\includegraphics[scale=0.14]{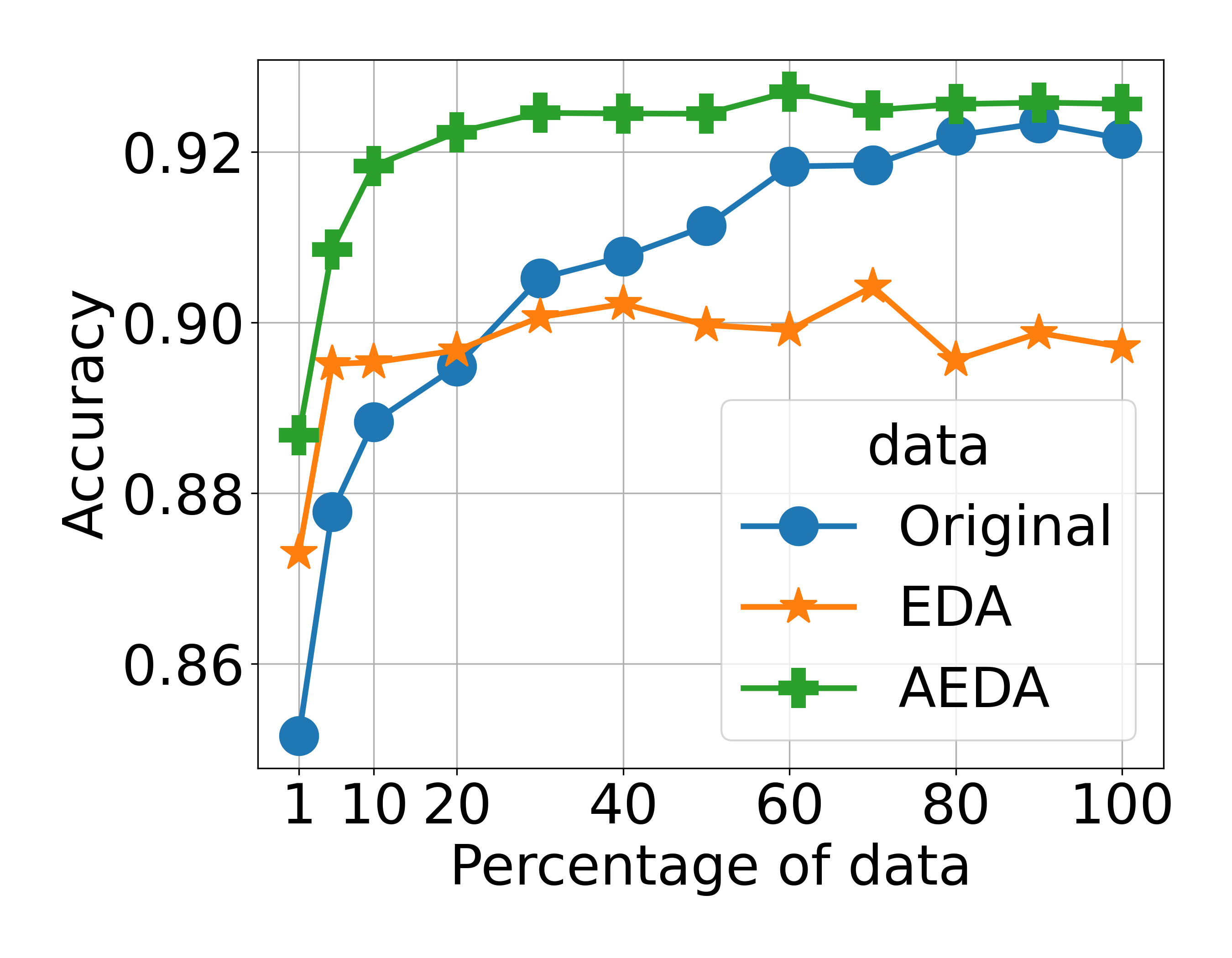}}
    \caption{Performance of the RNN model trained on various proportions of the original, EDA-generated, and AEDA-generated training data for five text classification tasks. All the scores are the average of 5 runs.}
    \label{fig:allfigures}
\end{figure*}

Figure \ref{fig:allfigures} shows how both models perform on different fractions of the training set. These fractions include \{1, 5, 10, 20, 30, 40, 50, 60, 70, 80, 90, 100\} percent. We can see that AEDA outperforms EDA in all tasks as well as showing improvements over the original data. One observation to point out is that also EDA works well on small datasets which can be because of lower number of augmentations compared to the ones reported in Table \ref{tab:resultsrnncnn}.

\begin{table}
    \centering
    \begin{tabular}{l|c c c c}
         &  \multicolumn{4}{c}{\textbf{Training set size}}\\
      \textbf{Model} & 500 & 2,000 & 5,000 & full set \\
      \hline
      RNN & 73.5 & 82.6 & 85.9 & 87.9 \\
      +EDA & 76.1 & 81.3 & 85.2 & 86.5\\
      +AEDA & 77.8 & 83.9 & 87.2 & 88.6\\
      \hline
      CNN & 76.5 & 83.8 & 87.0 & 87.9\\
      +EDA & 77.5 & 82.2 & 84.5 & 86.1\\
      +AEDA & 78.5 & 84.4 & 86.5 & 88.1\\
      \hline
      Average & 75.0 & 83.2 & 86.5 & 87.9 \\
      +EDA & 76.8 & 81.8 & 84.9 & 86.3\\
      +AEDA & \textbf{78.2} & \textbf{84.2} & \textbf{86.9} & \textbf{88.4}\\

    \end{tabular}
    \caption{Comparing average performance of EDA and AEDA across all datasets on different training set sizes. For each training sample, 16 augmented sentences were added. Scores are the average of 5 runs.}
    \label{tab:resultsrnncnn}
\end{table}

\section{Ablation Study}
In this section, we investigate how much gain there is for different number of augmentations, the effect of random initialization, and whether AEDA can improve deep models. 
\begin{figure}
    \centering
    \includegraphics[scale=0.3]{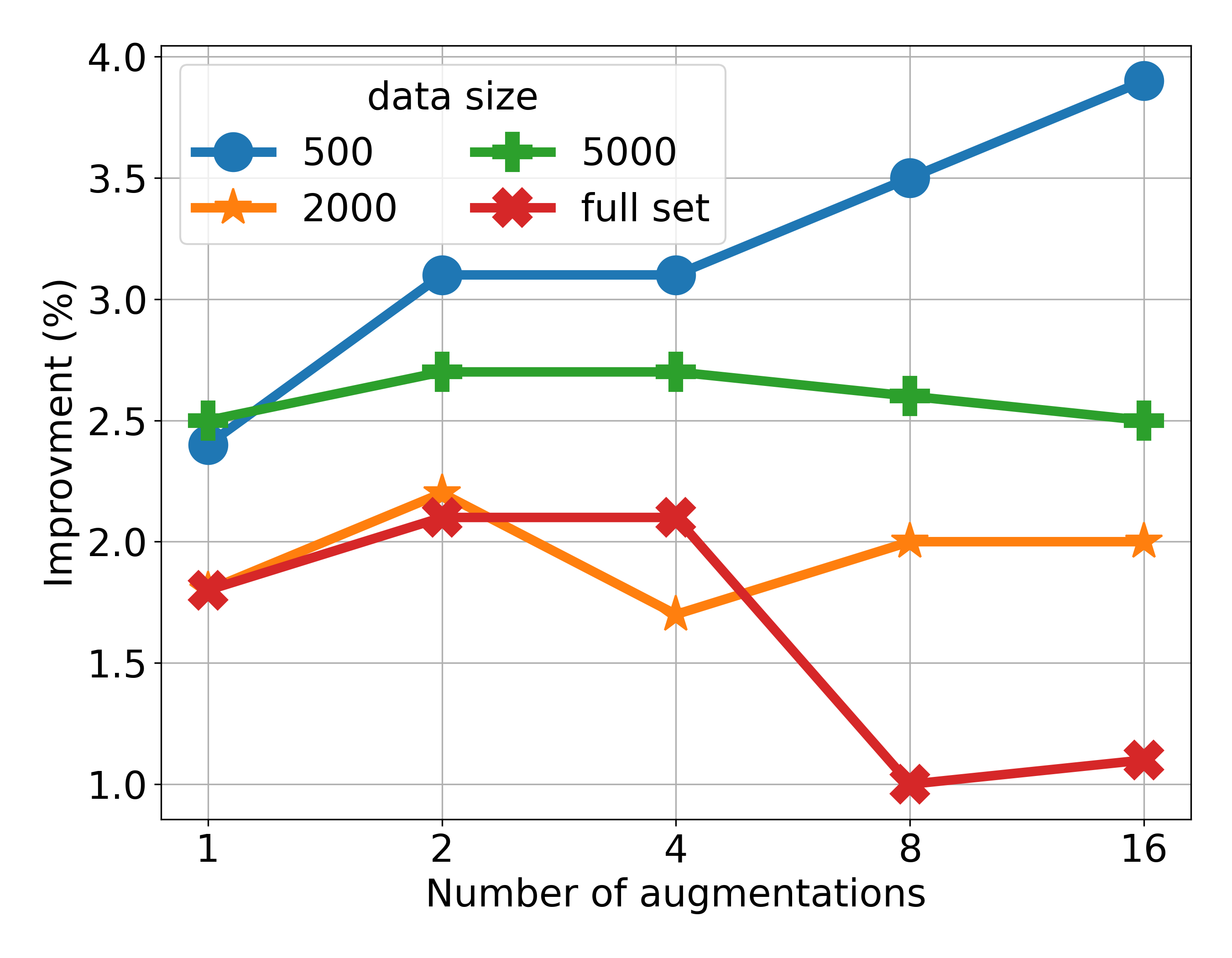}
    \caption{Impact of number of augmentations on the performance of the RNN model trained on various training sizes. Scores are the average of 5 runs over the five datasets. The y axis shows the percentage of improvement.}
    \label{fig:numaugs}
\end{figure}

\subsection{Number of Augmentations}
Figure \ref{fig:numaugs} presents the impact of adding various numbers of augmentations to the training set. We can see that only one augmentation can improve the performance by an absolute amount of 1.5\% to 2.5\% for all dataset sizes. However, as the augmentations increase, the smallest dataset greatly benefits from that by an improvement of almost 4\% while the full dataset only gains 1\%. The middle-sized ones have a gain in between (2\% to 2.5\%).

\subsection{Effect of Random Initialization}
\begin{figure}[h]
\subfloat[CR]{\includegraphics[scale=0.17]{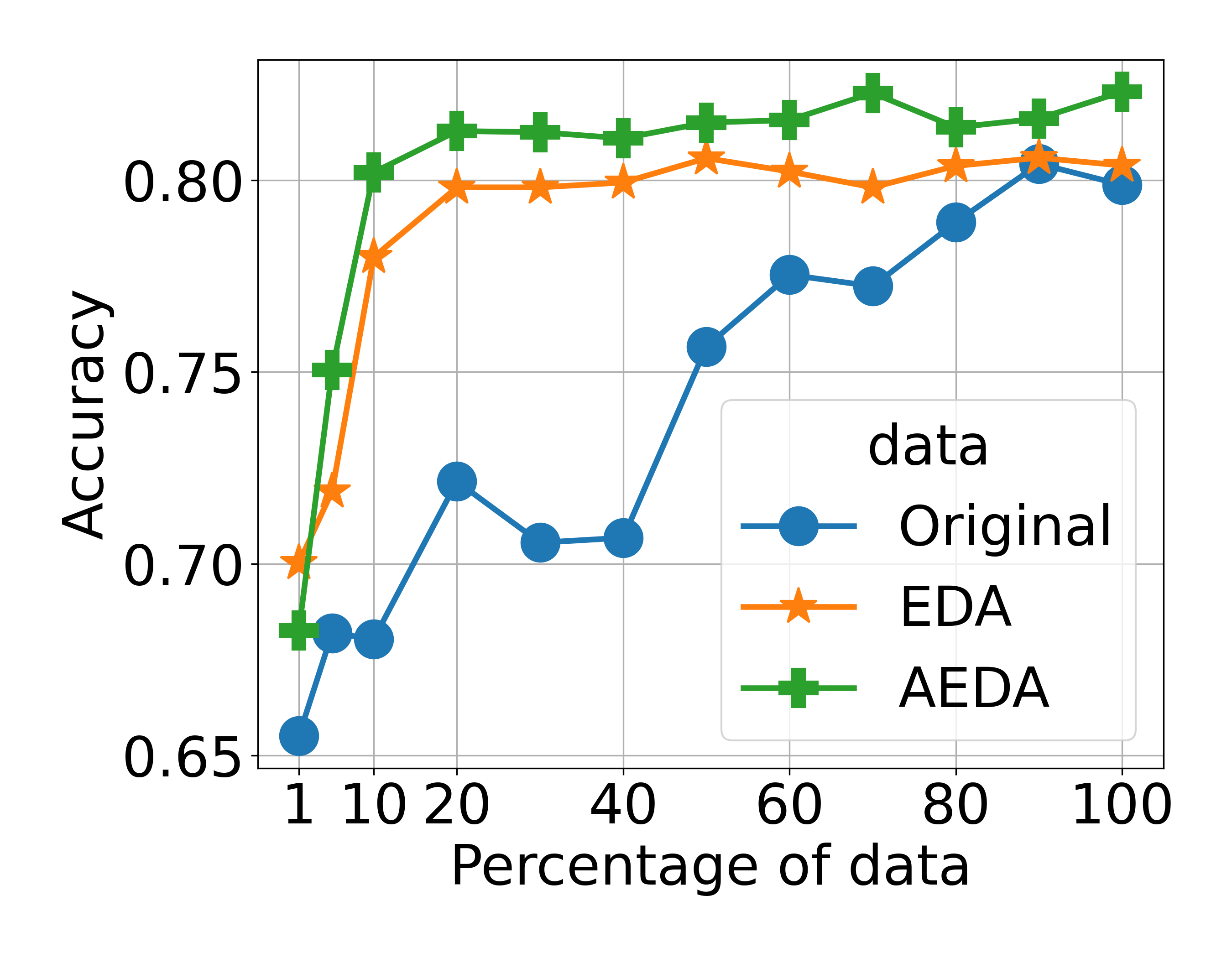}}
\subfloat[TREC]{\includegraphics[scale=0.17]{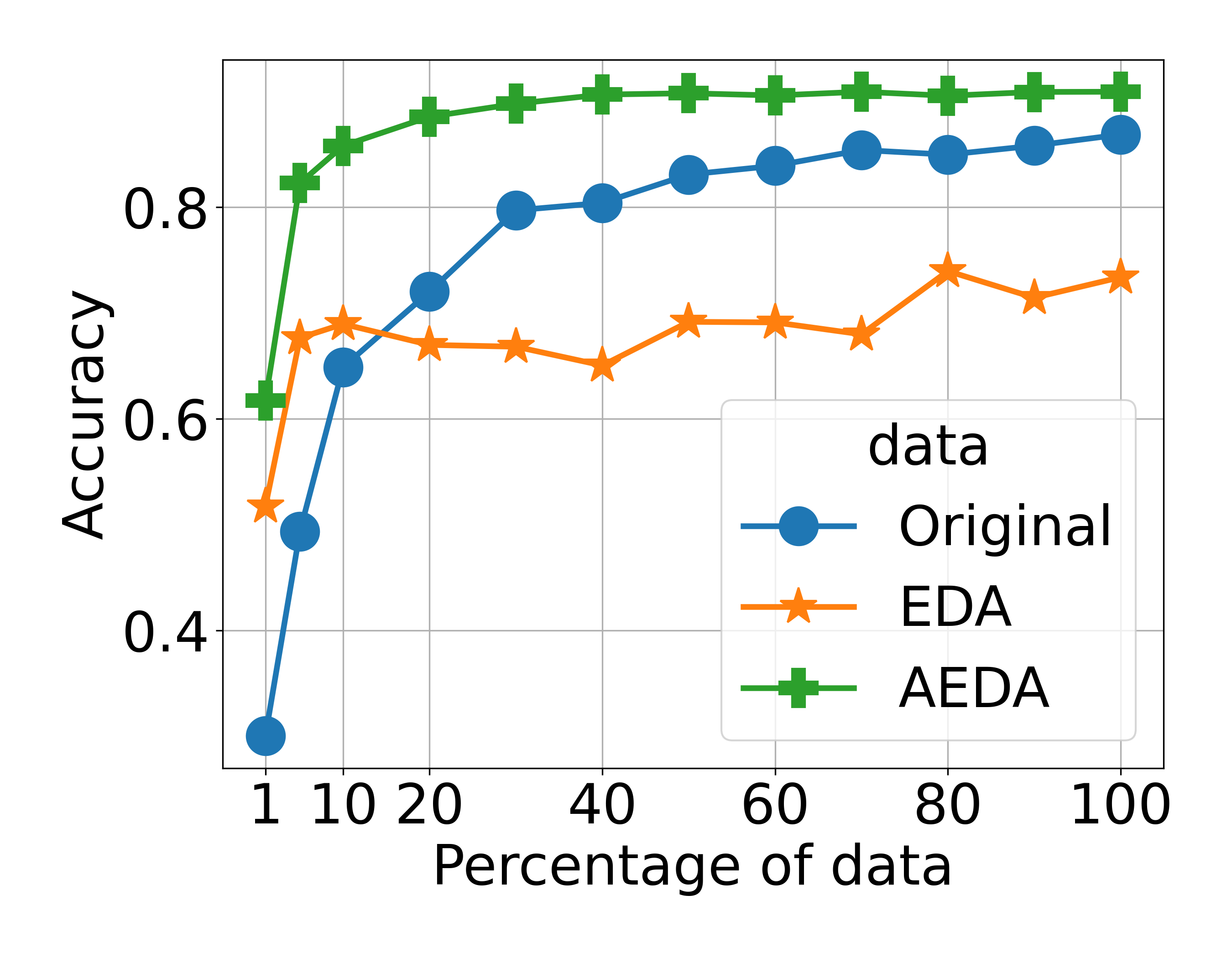}}
\caption{Average performance of EDA and AEDA over 21 different seed numbers. The results are in line with the experiments run over 5 seeds. }
\label{fig:fig21}
\end{figure}

When conducting the experiments, we noticed that different seed numbers produce different results. As a result, we ran the experiments for 5 times. However, in each run with the same seed number, the results can be slightly different due to the local and global generators in TensorFlow. Therefore, to ensure that 5 runs show the correct trend, we chose two of the datasets (CR and TREC) and ran the models for 21 different seeds (zero to 20). From Figure \ref{fig:fig21}, we see that the trend is similar to Figure \ref{fig:allfigures}, which shows the average results of 5 seeds.

\subsection{Using AEDA with Deep Models}
AEDA can also improve the performance of a deep model such as BERT. For instance, we trained the BERT model  used in \cite{kumar2020data} on SST2 and TREC for 3 epochs with its default settings and observed that adding one augmentation for each training sample increased the performance by 0.66\% for SST2 and 0.2\% for TREC (Table \ref{tab:bert}).

\begin{table}[h]
    \centering
    \begin{tabular}{l|c c}
        Model & SST2 & TREC \\
        \hline
        BERT & 91.10 & 97.00\\
        +EDA & 90.99 & 96.00 \\
        +AEDA & \textbf{91.76} & \textbf{97.20}\\
        
    \end{tabular}
    \caption{Comparing the impact of EDA and AEDA on the BERT model. The model was trained on the combination of the original data and one augmentation for each training sample. }
    \label{tab:bert}
\end{table}
\section{Discussion}

Comparing the results that we have gained in our experiments with the ones reported in \citet{wei2019eda}, we can see some discrepancy, especially in the impact of EDA on improving the performance of the models. We speculate that the difference can be caused by the inconsistency in the training and test sets. Although we obtained the datasets from the same references they have specified, some of them are not divided into train and test datasets ready to be used. As mentioned in Section \ref{stats}, we randomly divided them into train and test sets. In addition, some of them have different sizes which can produce different results.

With that said, to conduct a fair evaluation, we kept the same setting for all comparisons in terms of the utilized library and source code, train and test sets, number of augmentations, number of runs, batch size, and learning rate.

\section{Conclusion and Future Work}
We proposed an easy data augmentation technique for text classification tasks.  
Extensive experiments on five different datasets showed that this extremely simple method which uses punctuation marks outperforms the EDA technique which includes random deletion, insertion, and substitution of words, on all the utilized datasets. The future work will focus on exploiting the proposed method regarding which punctuation marks can have more impact, which ones to add or discard, and how many of them can be used to achieve a better performance. In addition, the question whether the punctuation marks should be inserted randomly or some positions are more effective will be investigated. 

\bibliographystyle{acl_natbib}
\bibliography{custom}

\begin{thebibliography}{29}
\expandafter\ifx\csname natexlab\endcsname\relax\def\natexlab#1{#1}\fi

\bibitem[{Andreas(2020)}]{andreas2020good}
Jacob Andreas. 2020.
\newblock Good-enough compositional data augmentation.
\newblock In \emph{Proceedings of the 58th Annual Meeting of the Association
  for Computational Linguistics}, pages 7556--7566.

\bibitem[{Devlin et~al.(2019)Devlin, Chang, Lee, and
  Toutanova}]{devlin2019bert}
Jacob Devlin, Ming-Wei Chang, Kenton Lee, and Kristina Toutanova. 2019.
\newblock Bert: Pre-training of deep bidirectional transformers for language
  understanding.
\newblock In \emph{Proceedings of the 2019 Conference of the North American
  Chapter of the Association for Computational Linguistics: Human Language
  Technologies, Volume 1 (Long and Short Papers)}, pages 4171--4186.

\bibitem[{Ding et~al.(2020)Ding, Liu, Bing, Kruengkrai, Nguyen, Joty, Si, and
  Miao}]{ding2020daga}
Bosheng Ding, Linlin Liu, Lidong Bing, Canasai Kruengkrai, Thien~Hai Nguyen,
  Shafiq Joty, Luo Si, and Chunyan Miao. 2020.
\newblock Daga: Data augmentation with a generation approach for low-resource
  tagging tasks.
\newblock \emph{arXiv preprint arXiv:2011.01549}.

\bibitem[{Ding et~al.(2008)Ding, Liu, and Yu}]{ding2008holistic}
Xiaowen Ding, Bing Liu, and Philip~S Yu. 2008.
\newblock A holistic lexicon-based approach to opinion mining.
\newblock In \emph{Proceedings of the 2008 international conference on web
  search and data mining}, pages 231--240.

\bibitem[{Fadaee et~al.(2017)Fadaee, Bisazza, and Monz}]{fadaee2017data}
Marzieh Fadaee, Arianna Bisazza, and Christof Monz. 2017.
\newblock Data augmentation for low-resource neural machine translation.
\newblock In \emph{Proceedings of the 55th Annual Meeting of the Association
  for Computational Linguistics (Volume 2: Short Papers)}, pages 567--573.

\bibitem[{Ganapathibhotla and Liu(2008)}]{ganapathibhotla2008mining}
Murthy Ganapathibhotla and Bing Liu. 2008.
\newblock Mining opinions in comparative sentences.
\newblock In \emph{Proceedings of the 22nd International Conference on
  Computational Linguistics (Coling 2008)}, pages 241--248.

\bibitem[{Garg and Ramakrishnan(2020)}]{garg2020bae}
Siddhant Garg and Goutham Ramakrishnan. 2020.
\newblock Bae: Bert-based adversarial examples for text classification.
\newblock In \emph{Proceedings of the 2020 Conference on Empirical Methods in
  Natural Language Processing (EMNLP)}, pages 6174--6181.

\bibitem[{Hu and Liu(2004)}]{hu2004mining}
Minqing Hu and Bing Liu. 2004.
\newblock Mining and summarizing customer reviews.
\newblock In \emph{Proceedings of the tenth ACM SIGKDD international conference
  on Knowledge discovery and data mining}, pages 168--177.

\bibitem[{Hu et~al.(2019)Hu, Tan, Salakhutdinov, Mitchell, and
  Xing}]{hu2019learning}
Zhiting Hu, Bowen Tan, Ruslan Salakhutdinov, Tom Mitchell, and Eric~P Xing.
  2019.
\newblock Learning data manipulation for augmentation and weighting.
\newblock \emph{arXiv preprint arXiv:1910.12795}.

\bibitem[{Jiao et~al.(2020)Jiao, Yin, Shang, Jiang, Chen, Li, Wang, and
  Liu}]{jiao2020tinybert}
Xiaoqi Jiao, Yichun Yin, Lifeng Shang, Xin Jiang, Xiao Chen, Linlin Li, Fang
  Wang, and Qun Liu. 2020.
\newblock Tinybert: Distilling bert for natural language understanding.
\newblock In \emph{Proceedings of the 2020 Conference on Empirical Methods in
  Natural Language Processing: Findings}, pages 4163--4174.

\bibitem[{Karimi et~al.(2020)Karimi, Rossi, Prati, and
  Full}]{karimi2020adversarial}
Akbar Karimi, Leonardo Rossi, Andrea Prati, and Katharina Full. 2020.
\newblock Adversarial training for aspect-based sentiment analysis with bert.
\newblock \emph{arXiv preprint arXiv:2001.11316}.

\bibitem[{Kim(2014)}]{kim2014convolutional}
Yoon Kim. 2014.
\newblock Convolutional neural networks for sentence classification.
\newblock In \emph{Proceedings of the 2014 Conference on Empirical Methods in
  Natural Language Processing (EMNLP)}, pages 1746--1751.

\bibitem[{Kobayashi(2018)}]{kobayashi2018contextual}
Sosuke Kobayashi. 2018.
\newblock Contextual augmentation: Data augmentation by words with paradigmatic
  relations.
\newblock In \emph{Proceedings of the 2018 Conference of the North American
  Chapter of the Association for Computational Linguistics: Human Language
  Technologies, Volume 2 (Short Papers)}, pages 452--457.

\bibitem[{Kumar et~al.(2020)Kumar, Choudhary, and Cho}]{kumar2020data}
Varun Kumar, Ashutosh Choudhary, and Eunah Cho. 2020.
\newblock Data augmentation using pre-trained transformer models.
\newblock In \emph{Proceedings of the 2nd Workshop on Life-long Learning for
  Spoken Language Systems}, pages 18--26.

\bibitem[{Li and Roth(2002)}]{li2002learning}
Xin Li and Dan Roth. 2002.
\newblock Learning question classifiers.
\newblock In \emph{COLING 2002: The 19th International Conference on
  Computational Linguistics}.

\bibitem[{Liu et~al.(2016)Liu, Qiu, and Huang}]{liu2016recurrent}
Pengfei Liu, Xipeng Qiu, and Xuanjing Huang. 2016.
\newblock Recurrent neural network for text classification with multi-task
  learning.
\newblock In \emph{Proceedings of the Twenty-Fifth International Joint
  Conference on Artificial Intelligence}, pages 2873--2879.

\bibitem[{Liu et~al.(2015)Liu, Gao, Liu, and Zhang}]{liu2015automated}
Qian Liu, Zhiqiang Gao, Bing Liu, and Yuanlin Zhang. 2015.
\newblock Automated rule selection for aspect extraction in opinion mining.
\newblock In \emph{Twenty-Fourth international joint conference on artificial
  intelligence}.

\bibitem[{Liu et~al.(2020)Liu, Xu, Jia, Ma, Wang, and Vosoughi}]{liu2020data}
Ruibo Liu, Guangxuan Xu, Chenyan Jia, Weicheng Ma, Lili Wang, and Soroush
  Vosoughi. 2020.
\newblock Data boost: Text data augmentation through reinforcement learning
  guided conditional generation.
\newblock In \emph{Proceedings of the 2020 Conference on Empirical Methods in
  Natural Language Processing (EMNLP)}, pages 9031--9041.

\bibitem[{Pang and Lee(2004)}]{pang2004sentimental}
Bo~Pang and Lillian Lee. 2004.
\newblock A sentimental education: Sentiment analysis using subjectivity
  summarization based on minimum cuts.
\newblock In \emph{Proceedings of the 42nd Annual Meeting of the Association
  for Computational Linguistics (ACL-04)}, pages 271--278.

\bibitem[{Ragni et~al.(2014)Ragni, Knill, Rath, and Gales}]{ragni2014data}
Anton Ragni, Kate~M Knill, Shakti~P Rath, and Mark~JF Gales. 2014.
\newblock Data augmentation for low resource languages.
\newblock In \emph{INTERSPEECH 2014: 15th Annual Conference of the
  International Speech Communication Association}, pages 810--814.
  International Speech Communication Association (ISCA).

\bibitem[{Sennrich et~al.(2016)Sennrich, Haddow, and
  Birch}]{sennrich2016improving}
Rico Sennrich, Barry Haddow, and Alexandra Birch. 2016.
\newblock Improving neural machine translation models with monolingual data.
\newblock In \emph{Proceedings of the 54th Annual Meeting of the Association
  for Computational Linguistics (Volume 1: Long Papers)}, pages 86--96.

\bibitem[{Socher et~al.(2013)Socher, Bauer, Manning, and
  Ng}]{socher2013parsing}
Richard Socher, John Bauer, Christopher~D Manning, and Andrew~Y Ng. 2013.
\newblock Parsing with compositional vector grammars.
\newblock In \emph{Proceedings of the 51st Annual Meeting of the Association
  for Computational Linguistics (Volume 1: Long Papers)}, pages 455--465.

\bibitem[{Sun et~al.(2020)Sun, Xia, Yin, Liang, Philip, and He}]{sun2020mixup}
Lichao Sun, Congying Xia, Wenpeng Yin, Tingting Liang, S~Yu Philip, and Lifang
  He. 2020.
\newblock Mixup-transformer: Dynamic data augmentation for nlp tasks.
\newblock In \emph{Proceedings of the 28th International Conference on
  Computational Linguistics}, pages 3436--3440.

\bibitem[{Vaswani et~al.(2017)Vaswani, Shazeer, Parmar, Uszkoreit, Jones,
  Gomez, Kaiser, and Polosukhin}]{vaswani2017attention}
Ashish Vaswani, Noam Shazeer, Niki Parmar, Jakob Uszkoreit, Llion Jones,
  Aidan~N Gomez, {\L}ukasz Kaiser, and Illia Polosukhin. 2017.
\newblock Attention is all you need.
\newblock In \emph{Proceedings of the 31st International Conference on Neural
  Information Processing Systems}, pages 6000--6010.

\bibitem[{Wang and Yang(2015)}]{wang2015s}
William~Yang Wang and Diyi Yang. 2015.
\newblock That’s so annoying!!!: A lexical and frame-semantic embedding based
  data augmentation approach to automatic categorization of annoying behaviors
  using\# petpeeve tweets.
\newblock In \emph{Proceedings of the 2015 Conference on Empirical Methods in
  Natural Language Processing}, pages 2557--2563.

\bibitem[{Wei and Zou(2019)}]{wei2019eda}
Jason Wei and Kai Zou. 2019.
\newblock Eda: Easy data augmentation techniques for boosting performance on
  text classification tasks.
\newblock In \emph{Proceedings of the 2019 Conference on Empirical Methods in
  Natural Language Processing and the 9th International Joint Conference on
  Natural Language Processing (EMNLP-IJCNLP)}, pages 6383--6389.

\bibitem[{Wu et~al.(2019)Wu, Lv, Zang, Han, and Hu}]{wu2019conditional}
Xing Wu, Shangwen Lv, Liangjun Zang, Jizhong Han, and Songlin Hu. 2019.
\newblock Conditional bert contextual augmentation.
\newblock In \emph{International Conference on Computational Science}, pages
  84--95. Springer.

\bibitem[{Xie et~al.(2019)Xie, Wang, Li, L{\'e}vy, Nie, Jurafsky, and
  Ng}]{xie2019data}
Ziang Xie, Sida~I Wang, Jiwei Li, Daniel L{\'e}vy, Aiming Nie, Dan Jurafsky,
  and Andrew~Y Ng. 2019.
\newblock Data noising as smoothing in neural network language models.
\newblock In \emph{5th International Conference on Learning Representations,
  ICLR 2017}.

\bibitem[{Zhang et~al.(2015)Zhang, Zhao, and LeCun}]{zhang2015character}
Xiang Zhang, Junbo Zhao, and Yann LeCun. 2015.
\newblock Character-level convolutional networks for text classification.
\newblock In \emph{Proceedings of the 28th International Conference on Neural
  Information Processing Systems-Volume 1}, pages 649--657.

\end{thebibliography}

\clearpage

\section{Supplementary Material}

\subsection{Example Augmentations}

\begin{table}[h]
\centering
    \begin{tabularx}{0.5\textwidth}{ l X }
    \hline
        \textbf{Original} & a sad , superior human comedy played out on the back roads of life . \\
        \hline
        \textbf{Aug 1} & a sad , superior human comedy played out on the back roads ; of life ; .\\
        \textbf{Aug 2} & a , sad . , superior human ; comedy . played . out on the back roads of life .\\
        \textbf{Aug 3} & : a sad ; , superior ! human : comedy , played out ? on the back roads of life .\\
        \hline
    \end{tabularx}
    \caption{Examples of augmented data using AEDA technique.}
    \label{tab:examples}
\end{table}
\subsection{Benchmark Datasets}

\begin{table}[h]
    \begin{tabularx}{\columnwidth}{l|X X X X X}
        Dataset & N$_{class}$ & L$_{avg}$ & N$_{train}$ & N$_{test}$ & |V| \\
        \hline
        SST-2 & 2 & 19 & 7791 & 1821 & 15771\\
        CR & 2 & 19 & 4067 & 451 & 9048\\
        SUBJ & 2 & 25 & 9000 & 1000 & 22715\\
        TREC & 6 & 10 & 5452 & 500 & 9448 \\
        PC & 2 & 7 & 40000 & 5806 & 26090\\
    \end{tabularx}
    \caption{Statistics of the utilized datasets. N$_{class}$: Number of classes, L$_{avg}$: Sentence average length, N$_{train}$: Number of training samples, N$_{test}$: Number of test samples, |V|: Number of unique words.}
    \label{tab:dataset}
\end{table}
\end{document}